\documentclass[conference]{IEEEtran}
\IEEEoverridecommandlockouts
\usepackage{cite}
\usepackage{amsmath,amssymb,amsfonts,bm}
\usepackage{algorithmic}
\usepackage{graphicx}
\usepackage{textcomp}
\usepackage{booktabs}
\usepackage{multirow}
\usepackage{xcolor}
\usepackage{tabularx}
\usepackage{pifont}
\def\BibTeX{{\rm B\kern-.05em{\sc i\kern-.025em b}\kern-.08em
 T\kern-.1667em\lower.7ex\hbox{E}\kern-.125emX}}

\begin{document}
\title{Adaptive Weighted Parameter Fusion with CLIP for Class-Incremental Learning}

\author{Juncen Guo\textsuperscript{1}, Xiaoguang Zhu\textsuperscript{2}, Liangyu Teng\textsuperscript{1}, Hao Yang\textsuperscript{1}, Jing Liu\textsuperscript{3}, Yang Liu\textsuperscript{4}$^\ast$, Liang Song\textsuperscript{1}$^\ast$
\\
\textsuperscript{1}Fudan University, 
\textsuperscript{2}University of California, Davis, 
\textsuperscript{3}The University of British Columbia,
\textsuperscript{4}Soochow University
}

\maketitle

\begin{abstract}
Class-incremental Learning (CIL) enables the model to incrementally absorb knowledge from new classes and build a generic classifier across all previously encountered classes. When the model optimizes with new classes, the knowledge of previous classes is inevitably erased, leading to catastrophic forgetting. Addressing this challenge requires making a trade-off between retaining old knowledge and accommodating new information. However, this balancing process often requires sacrificing some information, which can lead to a partial loss in the model's ability to discriminate between classes. To tackle this issue, we design the adaptive weighted parameter fusion with Contrastive Language-Image Pre-training (CLIP), which not only takes into account the variability of the data distribution of different tasks, but also retains all the effective information of the parameter matrix to the greatest extent. In addition, we introduce a balance factor that can balance the data distribution alignment and distinguishability of adjacent tasks. Experimental results on several traditional benchmarks validate the superiority of the proposed method.
 
\end{abstract}


\section{Introduction}
\label{sec:intro}

The training of deep neural networks generally relies on pre-collected datasets \cite{p1,p48}. However, in open-world scenarios, training data often arrive in stream format and cannot be stored long-term due to privacy and storage constraints \cite{p2}. This necessitates Class-Incremental Learning (CIL), which allows models to update with new classes incrementally while building a unified classifier for all encountered classes. A critical challenge in CIL is catastrophic forgetting, where training on new data overwrites previously learned knowledge, leading to irreversible performance degradation. Consequently, an effective CIL approach must address catastrophic forgetting while balancing old and new knowledge.

In the last decade, researchers have proposed various CIL methods to address this issue \cite{p4,p5,p6}. Specifically, data replay methods \cite{p4} retain and reuse a limited number of past examples, but are limited by storage limitations and potential privacy concerns. Data regularization methods employ previous data as reference metrics to guide model updates, while weight regularization methods selectively constrain parameter updates based on their importance to previous tasks \cite{p5}. %
Furthermore, knowledge distillation methods aim to preserve old knowledge by minimizing prediction discrepancies during incremental updates \cite{p5}. Despite these efforts, most methods face trade-offs between stability (retaining old knowledge) and plasticity (adapting to new tasks), often failing to achieve a satisfactory balance.

Another line of work explores parameter isolation, where subspaces of the parameter space are allocated to specific tasks \cite{p10,p11,p12,p13}. Fixed architecture approaches like HAT \cite{p10} and MEAT \cite{p11} use binary masks to freeze critical parameters for old tasks, while methods such as PackNet \cite{p13} and AGS-CL \cite{p14} dynamically identify and release fewer critical parameters for new tasks. Although effective in preventing forgetting, these methods often overlook the adaptability required to recognize new classes. Consequently, they struggle to maintain performance across both old and new tasks.

Traditional backbones for CIL include Multi-Layer Perceptrons (MLPs) \cite{p16}, Convolutional Neural Networks (CNNs) \cite{p17}, and Vision Transformers (ViTs) \cite{p18}. Although successful in many settings, these architectures are typically limited to visual features, restricting their application in complex, multimodal scenarios. Recently, pretrained visual language models such as CLIP \cite{p21} have demonstrated superior generalization capabilities, offering a robust foundation for downstream tasks\cite{p43,p45}. Methods leveraging these models, including prompt-based techniques \cite{p21} and adapters \cite{p24}, have shown promise in mitigating catastrophic forgetting with minimal parameter updates. For example, RARF \cite{p25} uses linear adaptive layers to integrate task-specific parameters and successfully reduces the foggeting of CLIP models. However, its reliance on manually set thresholds can limit both its flexibility and performance.

\begin{figure*}
   \centering
   \includegraphics[width=.83\textwidth]{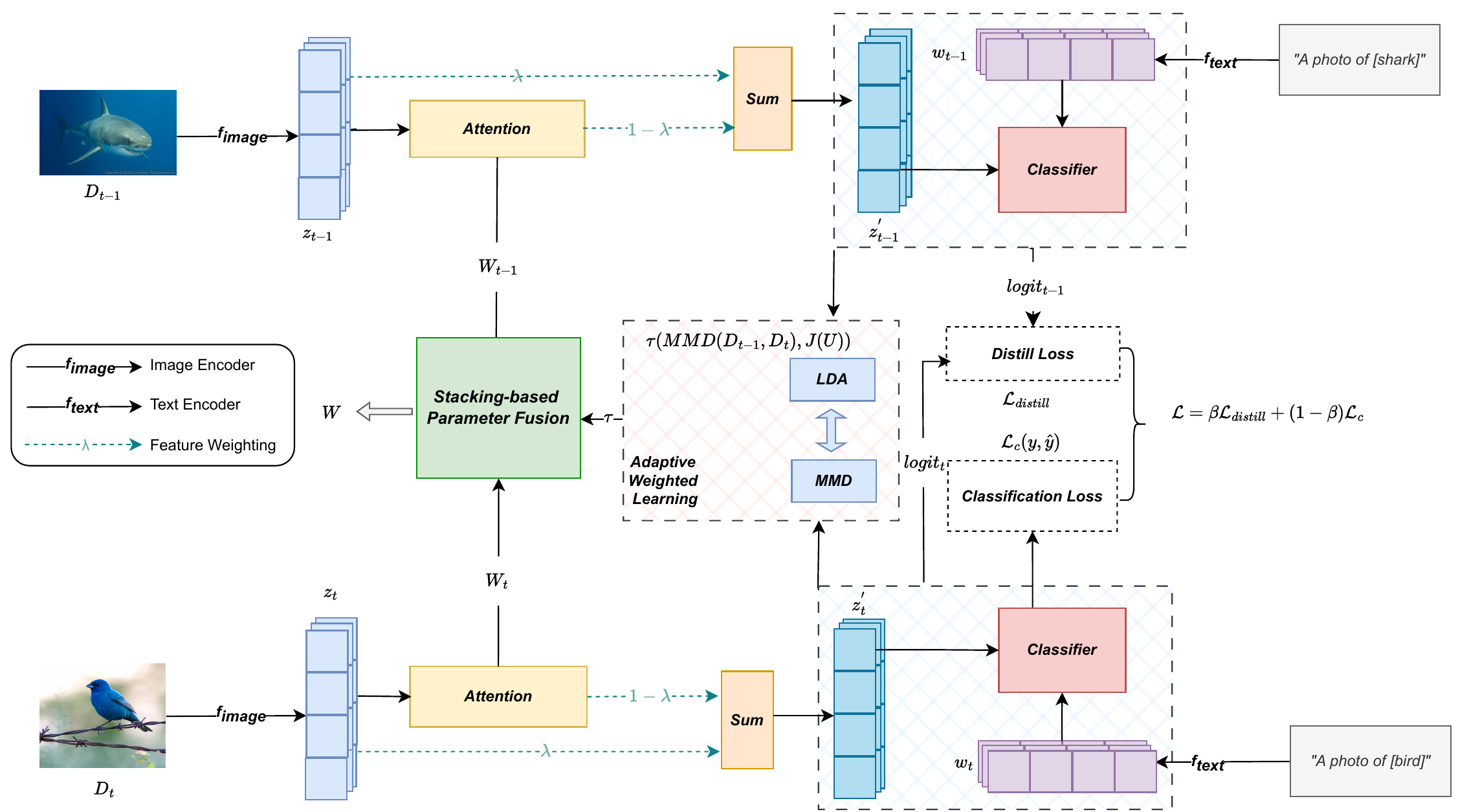}
   \caption{The framework of adaptive weighted parameter fusion with CLIP.}
   \label{fig1}
   \vspace{-12pt}
   \end{figure*}

To address these challenges, this paper proposes a novel CIL method based on adaptive weighted parameter fusion using the pretrained CLIP. The proposed approach employs the frozen CLIP encoder as a backbone, incorporating an adaptive parameter module for updates. This module retains the powerful generalization capabilities of the visual language model while adapting to new tasks with minimal overhead. A parameter fusion strategy based on low-rank decomposition and stacking is introduced to integrate task-specific parameters. This approach aligns parameter matrices across tasks, preserving adequate information and minimizing distributional shifts without sacrificing old or new knowledge. Notably, a dynamic balance factor is introduced, combining Maximum Mean Discrepancy (MMD) and Linear Discriminant Analysis (LDA) to dynamically adjust the fusion process, ensuring optimal trade-offs between task alignment and differentiation. Our contributions are summarized as follows:
\begin{itemize}
    \item We propose a novel parameter fusion method for CIL that effectively integrates knowledge retention and adaptation by leveraging low-rank decomposition and stacking.
    \item A dynamic balance factor combining MMD and LDA is introduced to adaptively regulate the parameter fusion process, addressing distributional variations within and across tasks.
    \item Comprehensive experiments on CIFAR100 and ImageNet100 demonstrate that our approach achieves state-of-the-art performance, highlighting its robustness and effectiveness in real-world incremental learning scenarios.
\end{itemize}

\section{Methodology}
\label{sec2}

\subsection{Overview}

The overall framework of our proposed method is shown in Fig.~\ref{fig1}, where data from two consecutive tasks arrive in sequence. Our framework is based on the frozen encoder of the pretrained visual language model CLIP, while adding adaptive parameter module. The module consists of a four-layer linear transformation based attention with feature weighting, and is the only component of the model that needs to be updated. 

For the parameter update mechanism of the adaptive parameter module, we design a stacking-based parameter fusion method. In this method, the parameter matrices of the adaptive module of different tasks are low-rank decomposed. The matrices after low-rank decomposition are aligned according to their respective dimensions, and the initial parameter matrix of the next task is obtained by multiplying the respectively aligned matrices. The stacking method not only considers that the tasks arriving in order have the difference in data distribution, but also retains all the effective information of the parameter matrix of the adjacent tasks to the greatest extent, which provides a new idea for the update and fusion of the old and new classes of knowledge. 

At the same time, in the process of stack fusion, we design the dynamic balance factor based on MMD and LDA. The balance factor takes into account the alignment and distinguishability of the data distribution of adjacent tasks, and further strengthens the adaptability of the model to the distribution differences within and between tasks. In addition to the initial task, the adaptive parameter module performs the computation of parameter fusion for the next round of tasks after learning each subsequent task in our method. Following the default method of CLIP, the final logits for classification are obtained based on the adaptive features and the textual features. The CLIP loss\cite{p21} is then computed and backpropagation is performed to update the parameters of the adaptive parameter module. Finally, we introduce the distillation loss\cite{p46}, then the new and old knowledge of the new and old tasks was effectively transferred and fused to alleviate the forgetting problem.

\subsection{Feature Optimization Adapter}
We add an additional adaptive parameter module to further improve the model as new data arrives. The module is divided into two parts. One part comprises four layers of linear transformations and incorporates the attention mechanism technique for feature enhancement. The other part is the weighted fusion of the enhanced features with the original CLIP image features. The above process is expressed as follows: 
\begin{equation}
   \textbf{z}_{t-1}^{\prime}=(1-\lambda) \phi_{f c}\left(\textbf{z}_{t-1}\right)+\lambda \textbf{z}_{t-1},
   \end{equation}
where $\lambda$ balances the enhanced features with the CLIP image features. The first term represents the enhanced features, and the second term is dedicated to preserving the original CLIP image features. The feature size before and after the adaptive parameter module is the same. The adaptive features can be combined with text features obtained after the CLIP text encoder. Following the default method of CLIP, the final logits for classification are obtained according to the adaptive features with the text features. The loss is then computed and backpropagation is performed to update the parameters of the adaptive parameter module. The adaptive parameter module is the only component that needs to be updated.

\subsection{Stacking-based Parameter Fusion}

In the class incremental learning scenario, each task contains different classes, and the data of different classes obviously have different distributions, that is, heterogeneity of data distribution. When tasks arrive sequentially, one by one, the distribution is unstable. And the model has to maintain good stability when dealing with data with unstable distribution. To this end, we design a parameter fusion method based stacking to fuse the distributional characteristics of different classes of adjacent tasks to minimize the impact of the unstable data distribution \cite{p26}. By low-rank decomposition of $\mathbf{W}_{}$, the most important information in $\mathbf{W}_{}$ is extracted, and then stacked and combined according to the main information, all the main information of $\mathbf{W}_{}$ obtained by two adjacent tasks is retained to the greatest extent. In addition, Low-Rank decomposition can also remove redundant data and noise interference \cite{p27}. 

As shown in Fig. 2, $\{\textbf{W}_{t-1}, \textbf{W}_{t} \} \in \mathbb{R}^{m \times n}
$ represents the adapter parameters between the previous task and the current task, respectively. $\textbf{W}_{t-1}$, $\textbf{W}_t$ first performs low-rank decomposition $\textbf{W}_{t-1}=\textbf{B}_{t-1}\textbf{A}_{t-1}$ and $\textbf{W}_t =\textbf{B}_t\textbf{A}_t$. Second, $\textbf{B}$ is obtained by stacking all $\mathbf{B}_{t}$ modules aligned with dimension $m$, and $\textbf{A}$ is obtained by stacking all $\mathbf{A}_{t}$ modules aligned with dimension $n$. Fig. 2 intuitively illustrates this concept, where the orange and blue rectangles represent $\mathbf{A}_{t}$, $\mathbf{B}_{t}$ and their respective products. The aggregation of two products mirrors the product of the stacked $\textbf{B}$ and $\textbf{A}$ from all $\mathbf{B}_{t}$ and $\mathbf{A}_{t}$ pairs. With the low-rank decomposition method, we mitigate the impact of differences in data distribution on model robustness by simply stacking the low-rank decompositions of the parameters of neighboring tasks. In other words, this method takes into account the effective integration of new and old knowledge simultaneously, without the need for difficult balance and trade-offs between new and old knowledge.

\begin{figure}[t]
   \centering
   \includegraphics[width=.49\textwidth]{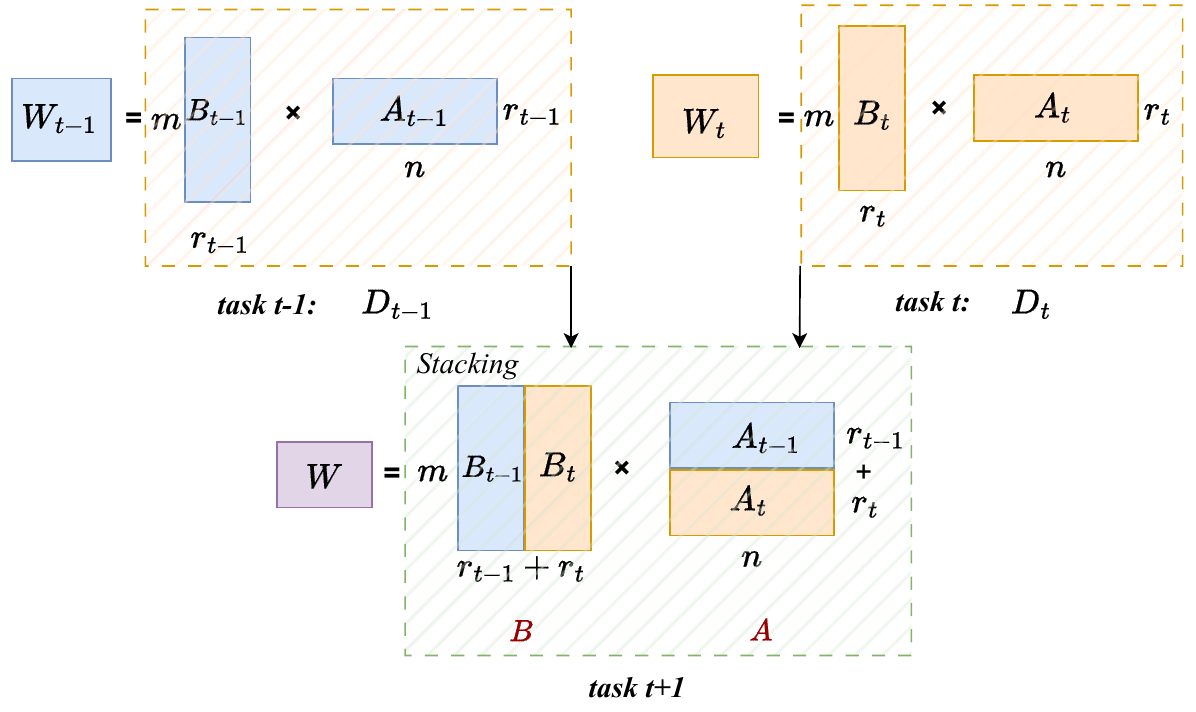}
   \caption{The process of parameter fusion method based on stacking.}
   \label{fig2}
   \vspace{-15pt}
\end{figure}

For the sake of discussion, we introduce the stacking operation denoted by $\oplus$ to represent module aggregation, as shown in Fig. 2. The stacking-based parameter fusion method is represented as $\mathrm{\textbf{A}}=\mathrm{\textbf{A}}_{t-1} \oplus \mathrm{\textbf{A}_{t}}, \quad \mathrm{\textbf{B}}=\mathrm{\textbf{B}}_{t-1} \oplus \mathrm{\textbf{B}_{t}}$, where $\oplus$ means that for $\textbf{A}$, the successor modules are stacked vertically below the previous module, while for $\textbf{B}$, the successor modules are stacked horizontally to the right of the previous module.

By adopting a stack-based aggregation mechanism, we design a dynamic and adaptive parameter fusion mechanism, which aims to promote the fusion of new and old category knowledge:
\begin{equation}
   \mathrm{\textbf{A}}=\tau \mathrm{\textbf{A}}_{t-1} \oplus \mathrm{\textbf{A}_t}, \quad \mathrm{~\textbf{B}}=\mathrm{\textbf{B}}_{t-1} \oplus \mathrm{\textbf{B}t},
   \end{equation}
where $\tau$ denotes the weight scaling factor for each update of $\textbf{W}$, which is calculated based on the class distribution of adjacent tasks. The details are described in Sec. \ref{sec}.

\subsection{Adaptive Weighted Learning}
\label{sec}
In the stack-based parameter fusion mechanism we designed, the stacking operation will amplify the value in $\mathbf{W}_{}$. We construct a dynamic balance factor $\tau$ to adaptively adjust the respective contributions of the new and old class distribution in $\mathbf{W}_{}$. MMD is used to calculate the degree of data distribution between adjacent tasks\cite{p28}, which is defined as follows:
\begin{equation}
\resizebox{0.85\linewidth}{!}{$
\text{MMD}(D_{t-1}, D_t) = \left\| \mathbb{E}_{x
 \sim D_{t-1}} G(\hat{x}_ { t-1 }) - \mathbb{E}_{x \sim D_t} G(\hat{x}_{ t }) \right\|^2
$},
\end{equation}
where $D_{t}$ denotes the training data of the t-th task. LDA is used to calculate the degree of distinguishability of the data distribution of adjacent tasks\cite{p28}. It is defined as follows:
\begin{equation}
   \max J(\textbf{U})=\frac { | \textbf{U} ^ { T } \textbf{S} _ { \textbf{b} } \textbf{U} | } { | \textbf{U} ^ { T } \textbf{S} _ { \textbf{w} } \textbf{U} | },
   \end{equation}
   where $\mathbf{\textbf{S}}_{\textbf{b}}$ is the between-class scatter matrix and $\mathbf{S}_{\textbf{w}}$ is the within-class scatter matrix\cite{p29}. $\mid \cdot \mid$ denotes determinant calculation. %
   Higher values of $J(\textbf{U})$ indicate better discrimination.

The MMD and LDA estimates are generally not of the same order of magnitude, and we normalize them separately using the min-max scaling method, which maps both values to the range [0, 1]. %
Based on the two normalized values, the dynamic balance factor $\tau$ is constructed, which is calculated as follows:
\begin{equation}
\tau = \frac{\text{MMD}'(D_{t-1}, D_t)}{\text{MMD}'(D_{t-1}, D_t) + (1 - J'(\textbf{U}))}.
\label{eq:q}
\end{equation}

In Eq.~\ref{eq:q}, a smaller MMD value indicates better distribution alignment and a smaller $1-J(\textbf{U})$ indicates better class discrimination. When the degree of distribution alignment is much better than class discriminability, MMD and $\tau$ are close to 0. Conversely, MMD and $\tau$ approach 1. When the degree of distribution alignment is equal to the degree of class discriminability, $\tau$ is approximately 0.5.

\section{Experiments}
\subsection{Experimental Setup}
\subsubsection{Datasets}
CIFAR100 \cite{p30} and ImageNet100 \cite{p38} are used for the dataset. The CIFAR100 dataset contains 100 classes. Each class contains 600 color images with a resolution of 32$\times$32 pixels. Each class contains 500 training samples and 100 testing samples. ImageNet100, containing samples with 224$\times$224 pixels from 100 classes. Each class contains about 1300 training samples and 50 testing samples. Our experimental setup unfolds according to the different number of tasks. That is, the total number of incremental learning phases T takes different values, which are 5 steps, 10th steps, and 20 steps. All classes are equally distributed across different tasks.

\subsubsection{Evaluation Metrics}
In the t-th incremental stage, the incremental accuracy refers to the classification accuracy At of the current model on all seen classes. In order to better compare the performance of different methods, the average incremental accuracy is defined as the average of the incremental accuracy of each incremental learning phase. $  Avg  = ( 1 / T ) \sum_{ i = 1 } ^ { T } A_{ i }$, where $T$ is the total number of incremental learning phases. Last is the average accuracy after the last task.

\subsubsection{Implementation Details}
We implemented the above methods on an NVIDIA RTX 4090 GPU using PyTorch. The backbone of CLIP uses the ViT-B/16 model, trained 15 times per task using the Adam optimizer, with an initial learning rate of 0.001, which is reduced by a factor of 0.1 at rounds 4 and 10. The batch sizes for ImageNet100 and CIFAR100 training are 128 and 100, respectively. Each task will simulate the replay samples by generating around 2000 samples from a Gaussian distribution of existing samples. The value of $\tau$ is 0.8. The text prompt format adopted by CLIP is "a good photo of [CLS]", where CLS is the name of the class to which the input data belongs.

\subsection{Comparison with Existing Methods}
We categorize the methods involved in the comparison into traditional and non-traditional methods. For fairness, we use the same CLIP pretrained weights for all non-traditional methods.

\subsubsection{Quantitative Results on Imagenet100}

\begin{table}[t]
   \centering
   \caption{RESULTS ON THE IMAGENET100 DATASET.}
   \label{tab1}
   \begin{tabular}{@{}lcccccc@{}}
   \toprule
   \multicolumn{1}{c}{\multirow{2}{*}{\textbf{Method}}} & \multicolumn{2}{c}{\textbf{20 steps}} & \multicolumn{2}{c}{\textbf{10 steps}} & \multicolumn{2}{c}{\textbf{5 steps}} \\ \cmidrule(l){2-7} 
   \multicolumn{1}{c}{}                                 & \textit{Avg}      & \textit{last}    & \textit{Avg}      & \textit{last}    & \textit{Avg}     & \textit{last}    \\ \midrule
   LUCIR \cite{p32}                                        & 64.70             & 47.80            & 70.50             & 55.30            & 76.00            & 64.00            \\
   End2End \cite{p33}                                     & 68.30             & 48.90            & 70.10             & 50.30            & 75.50            & 64.00            \\
   RM \cite{p34}                                        & 65.40             & 45.70            & 70.40             & 53.20            & 75.50            & 62.20            \\
   PODNet \cite{p35}                                       & 66.70             & 48.90            & 72.30             & 72.00            & 78.20            & 66.20            \\ \midrule
   DualPrompt \cite{p36}                                  & 75.40             & 61.10            & 80.65             & 67.38            & 84.65            & 74.24            \\
   L2P++ \cite{p37}                                       & 75.43             & 62.10            & 80.51             & 67.22            & 84.12            & 73.70            \\
   ADAM-Adapter \cite{p38}                                 & 85.78             & 75.72            & 85.84             & 76.40            & 85.85            & 77.08            \\
   PROOF \cite{p39}                                        & 86.92             & 75.52            & 84.71             & 72.48            & 81.92            & 68.56            \\
   RARF \cite{p25}                                        & 87.59             & \textbf{79.87}            & 87.51             & 80.23            & 86.72            & 80.10            \\ \midrule
   Ours                                                 & \textbf{88.04}    & 78.22   & \textbf{88.25}    & \textbf{80.34}   & \textbf{87.71}   & \textbf{81.04}   \\ \bottomrule
   \end{tabular}
   \vspace{-12pt}
\end{table}

Table~\ref{tab1} shows the results for the ImageNet100 dataset under three different experimental settings. It can be seen that the performance of the traditional methods is weaker overall than the method based on the visual language model. L2P++\cite{p37}, the weakest performer among the method based on the visual language model, performs better than PODNet\cite{p35}, the best performer among the traditional methods, by 8.73\%, 8.21\%, 5.92\% on AVG, respectively, under three different experimental settings. It shows that the CIL model using visual language model can achieve better results than the traditional basic skeleton. Using visual language model as the skeleton is already a good starting point, so the application of visual language model in the field of class incremental learning has great potential. Our method significantly outperforms methods based on visual language models. The accuracy of our method basically maintained around 88.00\%, while the accuracy of the last task also maintained around 80.00\%. The average accuracy of the five methods based on visual language model corresponding to three different experimental settings is maintained at 82.22\%, 83.84\% , 84.65\%, respectively, which is much weaker than our method. It shows that although our method stands at a relatively excellent starting point, the idea of the parameter fusion method based on stacking and introducing dynamic balance factor are scientific, reasonable and effective, which play a positive role in improving the performance of our model. In addition, our method is slightly better than RARF\cite{p25}, which also proposes a parameter fusion method with a fixed threshold. However, the adaptive dynamic balance factor is designed in our method, which should be the reason for the better performance of our method.

\begin{table}[]
   \centering
   \caption{RESULTS ON THE CIFAR100 DATASET.}
   \label{tab2}
   \begin{tabular}{@{}lcccccc@{}}
   \toprule
   \multicolumn{1}{c}{\multirow{2}{*}{\textbf{Method}}} & \multicolumn{2}{c}{\textbf{20 steps}} & \multicolumn{2}{c}{\textbf{10 steps}} & \multicolumn{2}{c}{\textbf{5 steps}} \\ \cmidrule(l){2-7} 
   \multicolumn{1}{c}{}                                 & \textit{Avg}      & \textit{last}    & \textit{Avg}      & \textit{last}    & \textit{Avg}     & \textit{last}    \\ \midrule
   LUCIR \cite{p32}                                       & 58.20             & 41.10            & 58.70             & 42.90            & 62.80            & 46.90            \\
   WA \cite{p41}                                           & 67.30             & 48.20            & 69.50             & 53.70            & 72.80            & 60.30            \\
   DER \cite{p30}                                        & 74.10             & 62.60            & 75.40             & 64.40            & 76.80            & 67.30            \\
   PODNet \cite{p35}                                       & 54.00             & 35.80            & 58.00             & 40.70            & 66.70            & 51.50            \\ \midrule
   DualPrompt \cite{p36}                                  & 79.74             & 69.91            & 81.45             & 72.51            & 85.19            & 77.47            \\
   L2P++ \cite{p37}                                       & 79.18             & 68.67            & 81.90             & 73.08            & 84.39            & 77.37            \\
   ADAM-Adapter \cite{p38}                                 & 70.18             & 58.12            & 80.53             & 65.50            & 77.28            & 67.89            \\
   Continual-CLIP \cite{p40}                               & 75.93             & 66.68            & 75.00             & 66.68            & 74.01            & 66.68            \\
   PROOF \cite{p39}                                        & 85.12             & 76.13            & 84.88             & 76.29            & 84.11            & 76.86            \\ \midrule
   Ours                                                 & \textbf{86.73}    & \textbf{79.00}   & \textbf{86.12}    & \textbf{79.46}   & \textbf{85.53}   & \textbf{79.84}   \\ \bottomrule
   \end{tabular}
   \vspace{-12pt}
\end{table}

\subsubsection{Quantitative Results on CIFAR100}

Table~\ref{tab2} shows the results for the CIFAR100 dataset under three different experimental settings. It can also be seen that the performance of the traditional method is weaker than that of the method based on visual language model. Continuous-CLIP\cite{p40}, which directly exploits the zero-shot performance of CLIP\cite{p42} for class incremental learning, performs almost on par with DER\cite{p30}, the best performing method among traditional methods. It shows that the class incremental model has stood at a relatively high starting point by using the visual language model, and can achieve better results than the traditional basic skeleton. Our method significantly outperforms methods based on visual language models. The accuracy of our methods basically stays around 86.00\%, and the accuracy of the last task stays around 79.00\%. The average accuracy of the five methods based on visual language model corresponding to three different experimental settings is maintained at 78.03\%, 80.75\%, 80.99\%, respectively. These methods are sensitive to different experimental settings, while our method is relatively stable. The accuracy of the last task of the five methods is also maintained at the level of 67.90\%, 70.81\%, and 73.25\%, respectively, which is also a gap from our method. It shows the effectiveness of our method. The parameter fusion method based on stacking and the idea of introducing a dynamic balance factor play a positive role in improving the performance of our model.

\subsection{Ablation studies}
\subsubsection{Ablation on Module}

\begin{table}[t]
   \centering
   \caption{ MODULE ABLATION RESULTS WITH IMAGENET100 B0 INC10.}
   \label{tab3}
   \begin{tabular}{@{}cccccc@{}}
   \toprule
   \textbf{FOA} & \textbf{SPF} & \textbf{AWL} & \textbf{DISTILL} & \textit{Avg}   & \textit{Last}  \\ \midrule
                &              &              &                  & 84.99          & 75.26          \\
   {\ding{51}}            &              &              &                  & 85.82          & 73.78          \\
   {\ding{51}}            & {\ding{51}}            &              &                  & 88.01          & 79.98          \\
   {\ding{51}}            & {\ding{51}}            & {\ding{51}}            &                  & 88.22          & 80.21          \\
   {\ding{51}}            & {\ding{51}}            & {\ding{51}}            & {\ding{51}}                & \textbf{88.25} & \textbf{80.34} \\ \bottomrule
   \end{tabular}
   \vspace{-15pt}
\end{table}

The results of ablation experiments based on our method are shown in Table~\ref{tab3}. The first row refers to the zero-shot performance of CLIP. Table~\ref{tab3} shows the results of different module ablation experiments for our proposed method. FOA brings 0.83\% improvement in terms of average accuracy, indicating that it is necessary to retain the original features of CLIP. However, it is more inclined to learn the knowledge of new classes, and it exacerbates the forgetting, which makes the last index decrease by 1.48\%.  SPF improves the average accuracy of the model by 2.19\%, and the last index by 6.2\%, indicating that the parameter fusion method based on stacking plays a positive role in both learning new knowledge and preventing the forgetting of old knowledge, which is consistent with our starting point of using the stacking method. The introduction of the dynamic balance factor further improves the performance of the model, which also verifies the rationality of our design idea of the dynamic factor. The introduction of distillation plays more of a role in solving the forgetting problem.
\begin{figure}[t]
   \centering
   \includegraphics[width=0.5\textwidth]{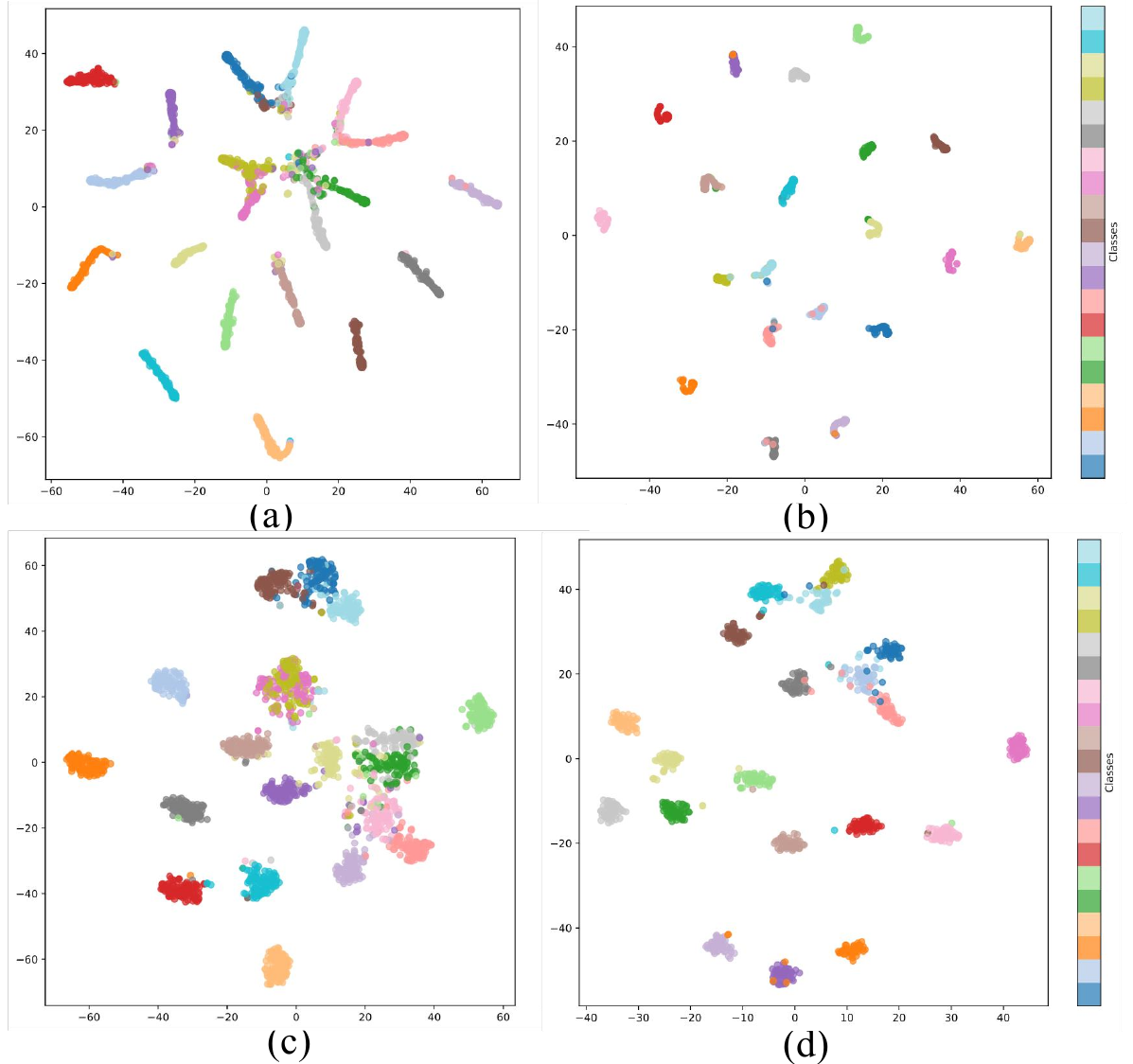}
   \caption{The T-SNE visualization of (a) CLIP with CIFAR100, (b) CLIP with ImageNet100, (c) Our method with CIFAR100, (d) Our method with ImageNet100 on B0 Inc10 after the second task}.
   \label{fig4}
   \vspace{-10pt}
\end{figure}
\subsubsection{Visual analysis of T-SNE}
Fig.~\ref{fig4} illustrates the T-SNE visualizations of different models applied to the CIFAR100 and ImageNet100 datasets. On the CIFAR100, the feature distribution produced by CLIP, which exhibits significant class overlap and dispersed intra-class features, indicating insufficient feature discrimination. In contrast, intra-class features are more compact, and inter-class separability is significantly improved, with distinct and independent clusters, highlighting the superior feature learning capability of our method. On the ImageNet100, the feature distribution generated by the CLIP model, showing slight improvements over CIFAR100 but still suffering from scattered clusters and overlapping classes. Meanwhile, for our method, the feature distributions of different classes are more cohesive and exhibit minimal overlap, demonstrating superior feature discrimination and robustness. Compared to the CLIP, our method consistently achieves better feature learning performance on the same dataset. It shows that the introduction of the stacking-based parameter fusion and the adaptive weighted learning is reasonable and effective.

\subsubsection{Ablation on Loss Functions}
As shown in Fig.~\ref{fig1}, we mainly use the classification loss and distillation loss, while introducing the parameter $\beta$ in the calculation of the loss function. As shown in Table~\ref{tab3}, the gain of the introduction of distillation loss is not obvious for the average accuracy of the model, which mainly improves the accuracy of the last task. When the distillation loss is introduced, the accuracy of the last task is improved by 0.13\%. In addition, we introduce $\beta$ as the proportion of old classes among all classes, which is dynamically adjusted, so that the loss function also realizes dynamic adaptation and achieves good results.

\subsubsection{Sensitivity Analysis on $\lambda$}

\begin{figure}[t]
   \centering
   \includegraphics[width=0.48\textwidth]{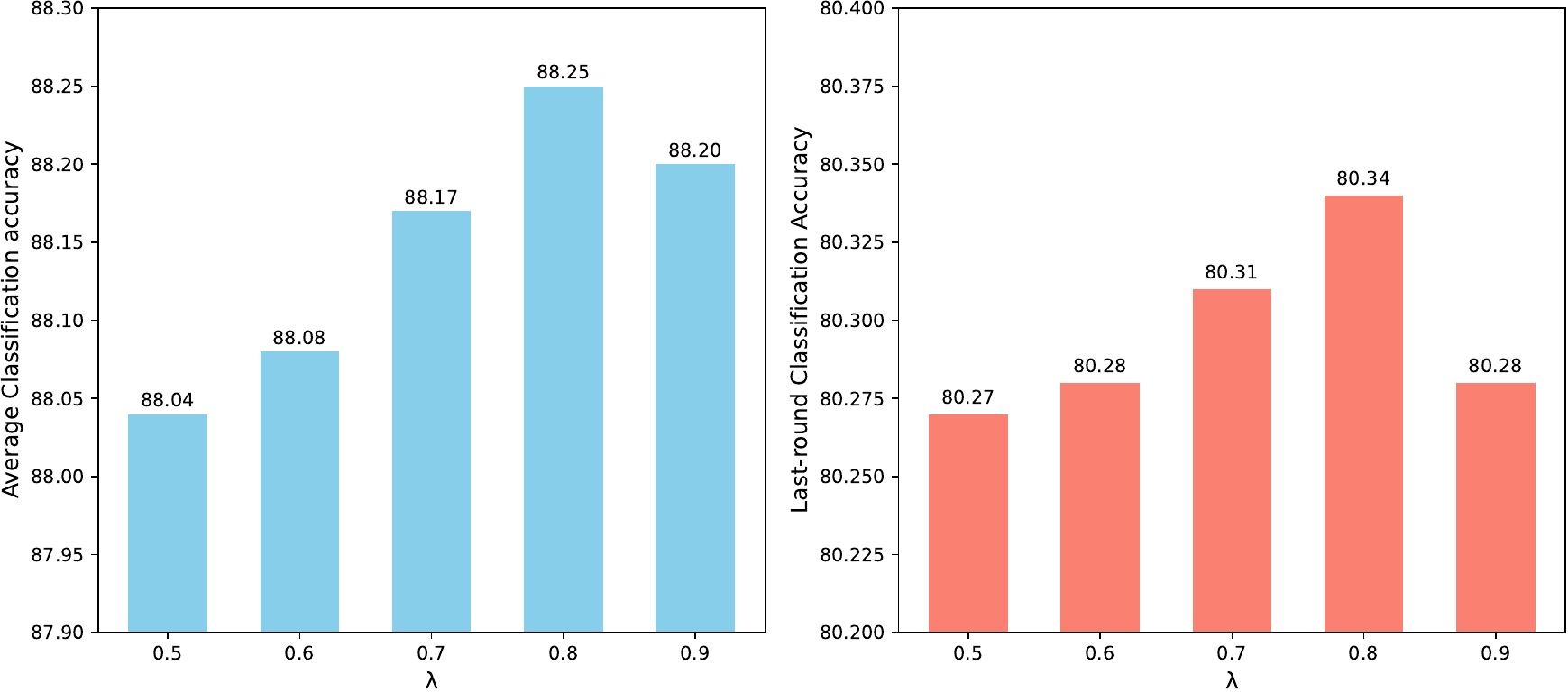}
   \caption{Sensitivity analysis on $\lambda$ with ImageNet100 B0 Inc10.}.
   \label{fig5}
\vspace{-18pt}
\end{figure}

When balancing the enhanced features with the CLIP image features, we introduce the parameter $\lambda$, which determines the retained ratio of the CLIP original image features. It can be seen from Fig.~\ref{fig5} that the value of $\lambda$ has an impact on the average accuracy of the model, indicating that whether and to what extent the original image features of CLIP are preserved should be concerned. Moreover, the size of $\lambda$ has less impact on the accuracy of the last task.

\section{Conclusion}
In this paper, we propose a novel class incremental learning method-adaptive weighted parameter fusion with CLIP. We propose a novel parameter update method for adaptive parameter modules -parameter matrix stacking method based on low-rank decomposition. The method does not need to make difficult trade-offs between the knowledge of new and old classes, and not only takes into account the difference of data distribution in different tasks, but also retains all the effective information of the parameter matrix to the greatest extent. At the same time, we design a dynamic balance factor based on MMD and LDA, which can take into account the data distribution alignment and discrimination of adjacent tasks, and further strengthen the adaptability of the model to the distribution differences within and between tasks. Experiments show the superiority of our method. Considering the limitation of new data acquisition in the open world, future work will focus on few-shot CIL.

\bibliographystyle{IEEEbib}
\bibliography{refs}

\end{document}